\documentclass{article}

\usepackage[final]{corl_2020} % Uncomment for the camera-ready ``final'' version.

\usepackage{amsmath}
\usepackage[pdftex]{graphicx}
\usepackage{subfig}
\usepackage{svg}
\usepackage{tabularx} 
\usepackage{booktabs}
\usepackage{gensymb}

\usepackage[font=footnotesize]{caption}
\usepackage{titlesec}
\titlespacing{\section}{0pt}{*0}{*0}
\titlespacing{\subsection}{0pt}{*0}{*0}
\newcommand{\etal}{\textit{et al.}}

\newcommand{\argmax}{\operatornamewithlimits{argmax}}

\title{Learning Arbitrary-Goal Fabric Folding with One Hour of Real Robot Experience\vspace{-0.4cm}}
\author{

  Robert Lee\\
  Queensland University of Technology \\
  \texttt{r21.lee@hdr.qut.edu.au } \\
  \And
  Daniel Ward \\
  CSIRO-Data61 \\
  \texttt{daniel.ward1@uq.net.au} \\
  \And
  Akansel Cosgun \\
  Monash University\\
  \texttt{akansel.cosgun@monash.edu}
  \And
  Vibhavari Dasagi \\
  Queensland University of Technology  \\
  \texttt{vibhavari.dasagi@hdr.qut.edu.au }
  \And
  Peter Corke \\
  Queensland University of Technology\\
  \texttt{peter.corke@qut.edu.au}
 \And
  J\"urgen Leitner \\
  Queensland University of Technology \\
  \texttt{juxi@lyro.io}
  \texttt{email} \\
}

\begin{document}
\maketitle
\vspace{-0.8cm}

%===============================================================================

\begin{abstract}
    Manipulating deformable objects, such as fabric, is a long standing problem in robotics, with state estimation and control posing a significant challenge for traditional methods. In this paper, we show that it is possible to learn fabric folding skills in only an hour of self-supervised real robot experience, without human supervision or simulation. Our approach relies on fully convolutional networks and the manipulation of visual inputs to exploit learned features, allowing us to create an expressive goal-conditioned pick and place policy that can be trained efficiently with real world robot data only. Folding skills are learned with only a sparse reward function and thus do not require reward function engineering, merely an image of the goal configuration.
    We demonstrate our method on a set of towel-folding tasks, and show that our approach is able to discover sequential folding strategies, purely from trial-and-error. We achieve state-of-the-art results without the need for demonstrations or simulation, used in prior approaches. Videos available at: https://sites.google.com/view/learningtofold
    
\end{abstract}

% Two or three meaningful keywords should be added here
% \keywords{Deformable Object Manipulation, Reinforcement Learning} 
\section{Introduction}
\label{sec:introduction}

The ability to interact with objects is crucial for building robotic systems that can work in the real world. The field of robotic manipulation has progressed significantly in recent years, in particular for tasks that require visual input, such as pick and place \cite{morrison2018cartman}, grasping \cite{morrison2020learning}, tossing \cite{zeng2019tossingbot}, and dexterous manipulation \cite{andrychowicz2020learning}. A major contributing factor is the success of learning-based approaches, in no small part due to the availability of large datasets, and the progression of computation and simulation. Despite recent advances, manipulation of deformable objects remains challenging as limited data exists for real-world interaction with deformable objects. Sim-to-real transfer is a promising direction for learning to solve complex robot tasks, yet requires a fast and accurate (enough) simulator. While there is progress, simulations of interactions with deformable objects are still slow and computationally expensive to create, due their incredibly complex dynamics.

\begin{figure}[t]
    \includegraphics[width=\linewidth]{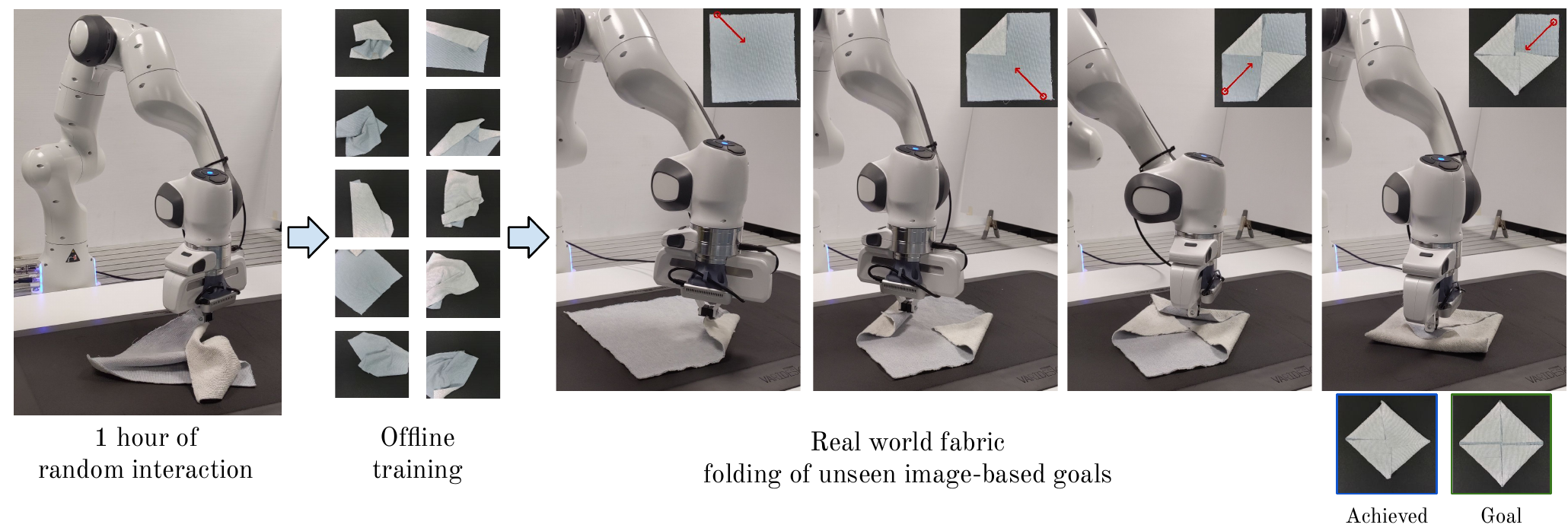}
    \caption{The full fabric folding system. First, we collect an hour of experience on the real robot, full self-supervised, with only random interactions with the fabric. Next, we train a fully-convolutional neural network agent on the fixed dataset. Finally, we deploy the trained agent on the robot with goals that were completely unseen during data collection and training.}
    \label{fig:hero}
\end{figure}

In this paper we present a method for a robot to learn to manipulate deformable objects directly in the real world, without reward function engineering, human supervision, human demonstration or simulation. We formulate the task of folding a piece of fabric into a goal configuration (Fig.~\ref{fig:hero}) as a reinforcement learning problem with top-down, discrete folding actions. We develop a method that is sample efficient, can learn effectively from sparse rewards, can achieve arbitrary unseen goal configurations given by a single image at test time. We also propose a novel self-supervised learning approach, in which the robot collects interaction data in the real world without human interaction. In particular, the contributions of this paper are:

\begin{itemize}
\setlength\itemsep{0pt}
    \item A fully-convolutional, deep Q-learning approach which leverages discretized folding distances for  sample-efficient real world learning. 
    \item A self-supervised learning pipeline for learning fabric manipulation via autonomous data collection by the robot in the real world.
\end{itemize}

\section{Related Work}
\label{sec:relatedwork}
\label{sec:related1}

Deformable object manipulation is a challenging area of robotics research \cite{sanchez2018robotic,cherubini2020model,staffa2015segmentation,arriola2017multimodal}. 
Fabric is a specific case that is of particular interest to robotics research, with a wide variety of real world applications. While methods exists that use specially designed tools for cloth manipulation \cite{osawa2006clothes}, in this work we are interested in manipulation with robot arms without significant modification to the end-effector. 

\label{sec:related2}

Cusumano~\etal \cite{cusumano2011bringing}, using a dual arm robot, demonstrated an approach that manipulates clothes to desired configurations via a Hidden Markov Model and a mesh simulation.

While the previous paper used a continuous action state space, others used discrete actions \cite{willimon2011model, sun2013heuristic, sun2015accurate}. Similarly, our work proposes a discrete action space for fabric manipulation, making learning behaviours more sample-efficient.

\label{sec:related3}
Recently, learning-based methods have dominated research in fabric manipulation, often using either human demonstrations \cite{laskey2017learning, nair2017combining, jangir2019dynamic}, sim-to-real transfer \cite{wu2019learning, seita2019deep, ganapathi2020learning, hoque2020visuospatial, yan2020learning} or both \cite{matas2018sim}.
Human demonstrations have been shown to improve  performance in fabric manipulation tasks \cite{matas2018sim, jangir2019dynamic}. Lasley~\etal \cite{laskey2017learning} used imitation learning for a bed-making task, where they showed that the trained policy had a comparable performance to the supervisor and outperformed a heuristic method based on contour detection. Jangir~\etal \cite{jangir2019dynamic} learned goal-conditioned reinforcement learning policies in simulation, with human demonstrations to improve learning performance. However, they utilized ground truth state information in sim, which is difficult to obtain in the real world. Our work therefore proposes to learn folding policies from real world vision only.
 
Matas~\etal \cite{matas2018sim} apply an end-to-end deep reinforcement learning approach, seeded with 20 human demonstrations in simulation and transfer the policy to real robot with domain randomization. In ablation studies, they show that human demonstrations were crucial in performance.

A recent trend in this area is that of training in simulation without human demonstrations and transferring the learned policies to real robots \cite{wu2019learning, seita2019deep, ganapathi2020learning, hoque2020visuospatial, yan2020learning}. Wu~\etal~\cite{wu2019learning}
learned continuous, folding policies efficiently in simulation without demonstration, by modelling the conditional relationship between pick and place actions. However, learning with this continuous action space in the real world is prohibitively expensive. In this work, we discretely model the action space to enable sample efficient learning of complex folding tasks in the real world.

Seita~\etal \cite{seita2019deep} framed the fabric flattening problem as imitation learning where an algorithmic supervisor
provides data in the form of paired observations.
They showed that using RGB as input transferred better than using only depth images. Our work utilizes only RGB images.
Ganapathi~\etal \cite{ganapathi2020learning} train a dense object descriptors model in simulation and showed successful transfer to two different robots in multi-step fabric smoothing and folding tasks. They collected task-agnostic data similar to our work, however their data was collected in simulation while we collect data directly on the real robot. Furthermore, at test time, they utilize human demonstrations to guide the policy through intermediate goals to solve sequential folding tasks. Our approach does not rely on demonstrations to solve such tasks.

Hoque \etal \cite{hoque2020visuospatial} learned an image prediction model in simulation, which can be used to solve arbitrary goals at test time via Model Predictive Control. They apply domain randomization to transfer fabric smoothing policies to a real-world surgical robot. However, their approach fails to transfer folding tasks successfully due to the sim-to-real gap.
 Yan~\etal \cite{yan2020learning} also used MPC along with model-based RL to transfer simulated policies to a real robot for cloth smoothing tasks. Both previous papers learn these models from offline, random actions in simulation. Our approach similarly uses random actions for data collection, but on a real robot to learn arbitrary folding tasks.

\section{Problem Formulation}
We propose to formulate the fabric manipulation problem as a Markov Decision Process (MDP), where an agent interacts with an environment by choosing an action $a_t$ from state $s_t$ according to the policy $\pi(s_t)$ from the full set of possible actions $a'$. The environment then transitions to a new state $s_{t+1}$ and the agent receives a reward according to reward function $R_{a_t}(s_t,s_{t+1})$. Reinforcement learning aims to obtain a policy that maximizes the expected sum of future rewards. We learn an state-action value function $Q(s_{t},a_{t})$, which predicts the expected sum of discounted future rewards after taking action $a_{t}$ from state $s_{t}$. We can optimize this function via the typical Q-learning formulation, i.e. minimizing the temporal difference error between $Q(s_{t},a_{t})$ and a target value $y_{t}$, estimated from data collected by the agent:
\begin{equation}
  y_{t} = R_{a_t}(s_t,s_{t+1}) + \gamma Q(s_{t+1}, \argmax_{a'}(Q(s_{t+1}, a')))
  \label{eq:qtarget}
\end{equation}
Where $\gamma$ is the discount factor.
We evaluate the greedy policy by choosing the action that maximises the predicted state-action value:
\begin{equation}
a_t = \argmax_{a'}(Q(s_{t+1}, a'))
\label{eq:greedy}
\end{equation}

\section{Approach}
\label{sec:approach}
\vspace{-0.1cm}

In this section we describe our method for learning fabric folding skills directly in the real world, with no reward function engineering, supervision, demonstration or simulation. In order to achieve this, we develop a method that is sample efficient, can learn effectively from sparse rewards, can achieve arbitrary unseen goal configurations given by a single image at test time, and requires no human interaction during data collection.

Our approach involves a two-stage process: data collection and offline training. The first step is autonomous data collection which occurs once. Our fabric folding policy is then learned via offline reinforcement learning.

\subsection{Action and Observation Spaces}
At each timestep, the agent observes a single RGB image captured by a downward-facing wrist-mounted camera, capturing the robot's effective workspace. Due to the complex dynamics involved in fabric manipulation, the choice of action space is important for learning efficiency. We use a discrete action space that consists of pixel grasp locations and predefined sets of fold angles and fold distances. The discrete action space, combined with our fully convolutional state-action value function, enables us to learn folding behaviour without simulation and a limited amount of offline data from a real robot.

\subsection{Data Collection}

We train our policy on a data-set of real robot experience, collected from random interactions. Actions are selected by choosing a pixel on the fabric uniformly at random, as well as a discrete distance to fold. 
We bias the direction of our random actions to move towards the centre of the workspace.
However we note that as the pixel location and distance is chosen uniformly at random, the actions do not result in moving the fabric directly to the centre, as it will often move past the centre to the other side of the workspace. Therefore, this bias prevents the random actions from pulling the fabric out of view, and continually moves the fabric around the workspace. We note that it does not bias the training data, as the network itself does not observe the angle of the fold, but rather the image that is rotated to the corresponding angle.

A wide distribution of experience is obtained during data collection by performing a soft-reset at regular intervals. This involves grasping the centre of the fabric and dropping it from a fixed height. Randomness is introduced as the fabric lands differently each time.

We collect the experience in tuples of $o_t$, the observation at time $t$, $a_t$, the action at time $t$, and $o_{t+1}$, the observation immediately following this action. This data is then used for offline training of the Q-network.

\subsection{State-Action Value Function}

We parameterize our action-value function as a feed-forward fully convolutional network, inspired by Zeng~\etal~\cite{zeng2018learning, zeng2019tossingbot} which learns heatmaps of visual-affordances over pixels with a network trained to evaluate actions such as grasping and fixed distance pushing from a single orientation, with input images being rotated to achieve a discrete set of possible actions. This, in combination with the weight sharing properties and spatial preservation of fully convolutional networks, makes this approach highly sample efficient. We apply this approach to fabric manipulation, while expanding the action space to variable distance folds, by exploiting the proportionality of scale when folding fabric.

The input to this network is the current observation (RGB image) of the state, channel-wise concatenated with the goal image. The network's output is a heatmap of the same resolution as the input image, representing the Q-value for a unit action performed at each pixel in the input image. 
We rotate the image to a canonical orientation prior to estimating the heatmap. Unit actions exist in this canonical orientation. In this way, we can generate heatmaps for arbitrary orientations by transforming the input image, and thus cover the full range of rotation by discretizing into a fixed number of fold angle bins.

\begin{figure}[b!]
\centering
\includegraphics[width=0.7\textwidth]{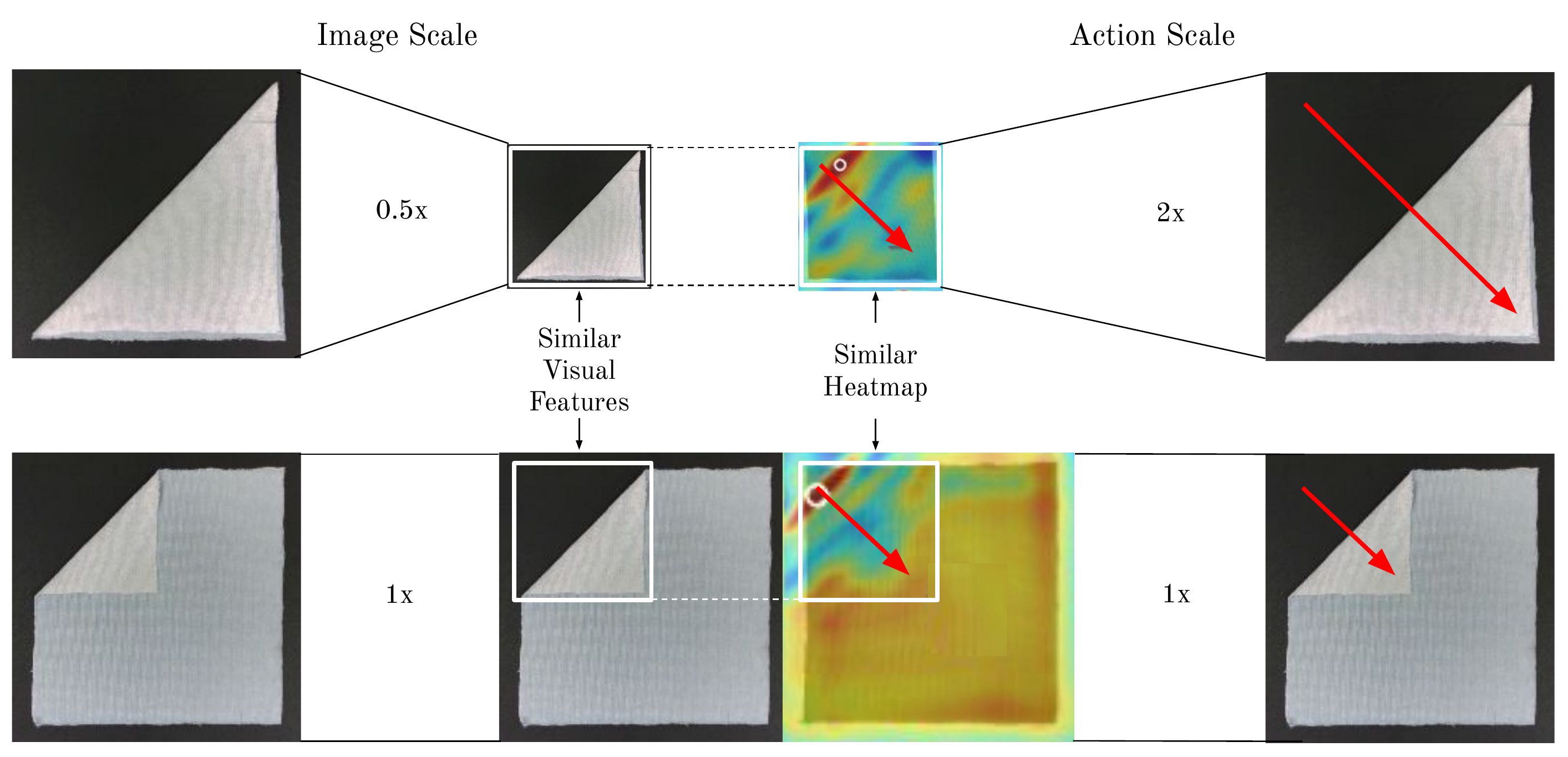}
\caption{Image scaling is utilized to produce varying fold distances, using a single trained CNN. Note the visual similarity of the regions of the image highlighted in white. The fold on the left is twice the size of that on the right, and so when scaled they appear visually similar. Thus, in the downscaled image on the left, the heatmap produced closely resembles the corresponding region of the heatmap on the bottom right.}
\label{fig:proportionalfeatures}
\end{figure}

However, a variable folding distance between pick and place is required in order to perform a range of folding tasks. We make the assumption that the visual appearance of folds in fabric scale proportionally with the length of the fold action. To this end, we extend the prior approach by additionally scaling the input image into the reference frame of the unit action, resulting in a heatmap for actions performed with distance relative to that image. The unit action produced by the network is then scaled by the reverse of the image scaling. If an input image is scaled by half, the corresponding output heatmap now represents folds that are twice the distance when executed on the real system. This is visualized in Figure \ref{fig:proportionalfeatures}. The relationship between action and image scale is:

\begin{equation}
    D_a = \frac{1}{\beta}D_u
\end{equation}

Where $D_a$ is the resulting fold distance, $\beta$ is the image scale factor, and $D_u$ is the unit fold distance.

\begin{figure}[t]

\centering
\includegraphics[width=0.8\textwidth]{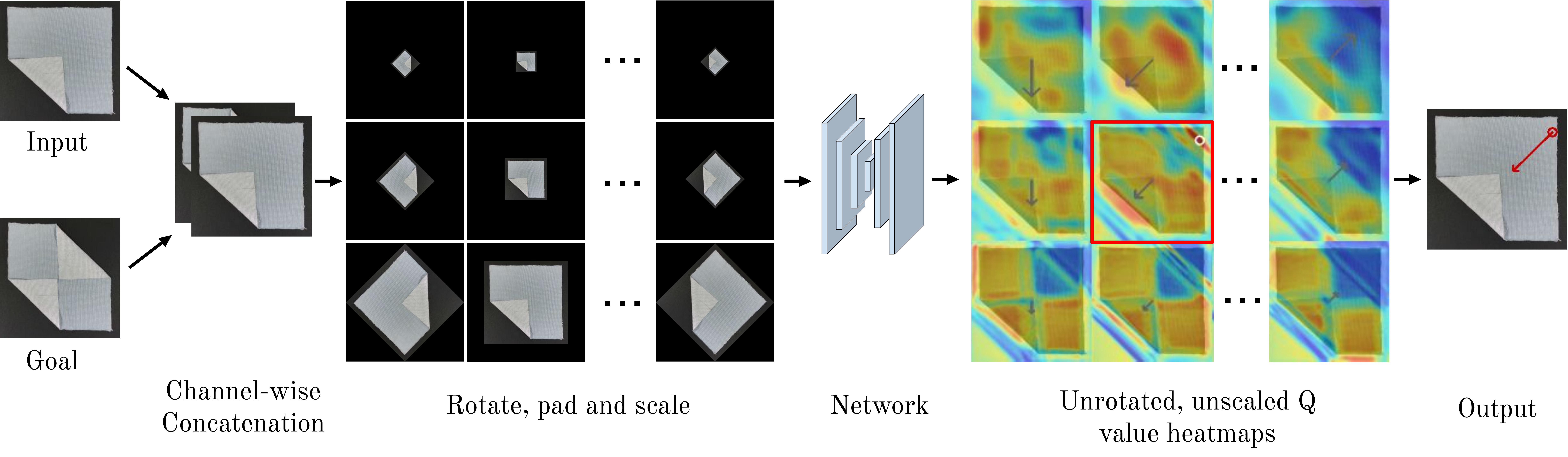}
\caption{Overview of our approach. We create an expressive fabric folding policy by manipulating the input image. We scale and rotate the image and pick the max value over the output heatmaps}
\label{fig:approachNetwork}

\end{figure}

To retrieve the next action from this pipeline, we define a fixed number of rotation bins 
and distance bins.
As shown in Figure \ref{fig:approachNetwork}, the input RGB image is rotated and scaled, and these images are passed through the network. The action is selected using Equation \ref{eq:greedy}, resulting in a pixel location, and an index of rotation and scaling.

\subsection{Offline Training}
The term Batch Reinforcement Learning (Batch RL) is used for learning policies from a fixed-batch of previously collected experience data. This is of particular interest to the robotics community, where real world data collection is costly. In the continuous domain particularly, off-policy reinforcement learning algorithms \cite{fujimoto2019off} fail to directly learn from fixed, offline datasets. However, successful batch learning in the discrete setting has been reported \cite{agarwal2020striving}. 

We use Batch RL for this work: once the robot has collected the dataset of random experience, we utilize this data to train our Q-network in an offline manner \cite{fujimoto2019off,agarwal2020striving, cabi2019scaling}. In order to learn a goal conditioned Q-function from sparse rewards, we utilize Hindsight Experience Replay~\cite{andrychowicz2017hindsight}. In our setting, when computing the Q-target value (Equation \ref{eq:qtarget}), we sample experience tuples $(o_t, a_t, o_{t+1})$ from the dataset and randomly select either an additional observation from the dataset as a goal, or the achieved next observation, with equal probability. We label our transitions with rewards according to this goal. As our reward is sparse, the reward is 1 when the sampled goal is the next achieved observation, and 0 otherwise.

In order to prevent overfitting in the low data regime and improve robustness of our learned policy, we employ simple data augmentation techniques to widen the distribution of training examples. While aiming to reach a goal configuration presented as an image, we would like to reduce sensitivity to slight translations and misalignments of the fabric in the image. To this end, we apply a small amount uniform random translation and rotation noise to observations sampled from the dataset for training. Similar data augmentation techniques have been utilized successfully for learning real robot skills from vision \cite{bruce2018learning}.

Batch RL is typically associated with instability issues. Many of the typical issues with fixed batch reinforcement learning arise more prominently in the continuous case \cite{agarwal2020striving, fujimoto2019off, siegel2020keep}. This further motivates our choice to discretize the fabric folding action space. However, training on offline data can still be a challenge, as the agent is never allowed to collect further evidence for the function it has approximated, and the distribution of training data might differ from the distribution of states the trained policy might experience. However, the difficulties of such offline training methods is reduced in our case. Despite the true action-space of the task consisting of pixel location, discrete angle and distance, the network does not observe the latter two. It only encodes a single function, the value of folding each pixel a fixed distance to the right. This is further simplified by the weight sharing properties of convolutional networks, where kernels are moved across the image. Thus, the training data is not required to exhaustively cover the state-action space, rather, information learned from a particular location in the canonical orientation, transfers immediately to anywhere in the image and to any fold angle or distance.

\section{Experiments}
\vspace{-0.1cm}

\label{sec:results}

In this section we describe the experimental evaluation of our method and discuss the results. Assessed quantitatively and qualitatively, we demonstrate the performance of our model on six different folding tasks. Finally, the ability of our approach to generalize to higher resolution discrete action-spaces and the limitations of our approach are explored.

\subsection{Experimental Setup}

We use the Franka Emika Panda robot arm with standard gripper, equipped with an eye-in-hand RGB camera, for data collection and testing. The image has a resolution of 200x200 pixels, which, with a camera height of 45cm, corresponds to a square workspace of 0.33x0.33 meters. 

Our fabric is a square, 30cm tea towel. The folding actions are performed on the robot via motion planning. Using simple force sensing, the robot attempts a top-down grasp at the 3D point corresponding to the selected pixel, and then performs a straight line fold in the specified direction and distance, before lowering and dropping the fabric.

 The action-space is discretized as a pixel grasp location, as well as a discrete set of rotation and distance options. During data collection and training, we utilize discrete the angles into 8, 45\degree bins. With the unit action in our approach being defined as 13cm (corresponding to the original size), the image is scaled by a factor of 2x, 1x and 0.5x before being passed as input into the CNN, resulting in three corresponding action bins of 6.5cm, 13cm, and 26cm (the largest covering almost the full width of the fabric). However, as explored in section \ref{sec:gen}, we evaluate the ability of our method to act (without further training) with a larger, higher resolution set of discretized actions. In this case, the angles are discretized into 16, 22.5\degree bins. The images are scaled by 2x, 1.5x, 1x, 0.66x and 0.5x, resulting in action bins of 6.5cm, 8.6cm, 13cm, 19.5cm, 26cm. As in \cite{wu2019learning}, when acting, we color-mask the fabric and choose pick locations only within the mask.

We utilize a fully-convolutional encoder of 4 layers (32 filters of size 5, stride of 2 for the first three layers followed by stride of 1 for the final layer), and construct the heatmaps by interleaving 2 convolutional layers (32 filters of size 3, stride of 1) with bilinear upsampling. We train our model with standard DQN \cite{mnih2015human}. Following Zeng~\etal \cite{zeng2018learning, zeng2019tossingbot}, and because the folding tasks we explore can be achieved in relatively few actions, we use a $\gamma$ of 0.5, and the Q-learning loss for each action is propagated through the single pixel corresponding the selected action. 

We collect 300 samples of training data, which with an average of 12 seconds per action is equivalent to 60 minutes of data collection time on the real robot. We use Huber loss \cite{huber1964} for our loss function, with a learning rate of 0.0001 and a batch size of 10. Early stopping is used to end offline training, when the loss consistently drops below an experimentally derived threshold of 0.05, which occurs after approximately 20,000 gradient steps.

\subsection{Folding Tasks}
\begin{figure}[tbh]
  \centering
  \includegraphics[width=\textwidth]{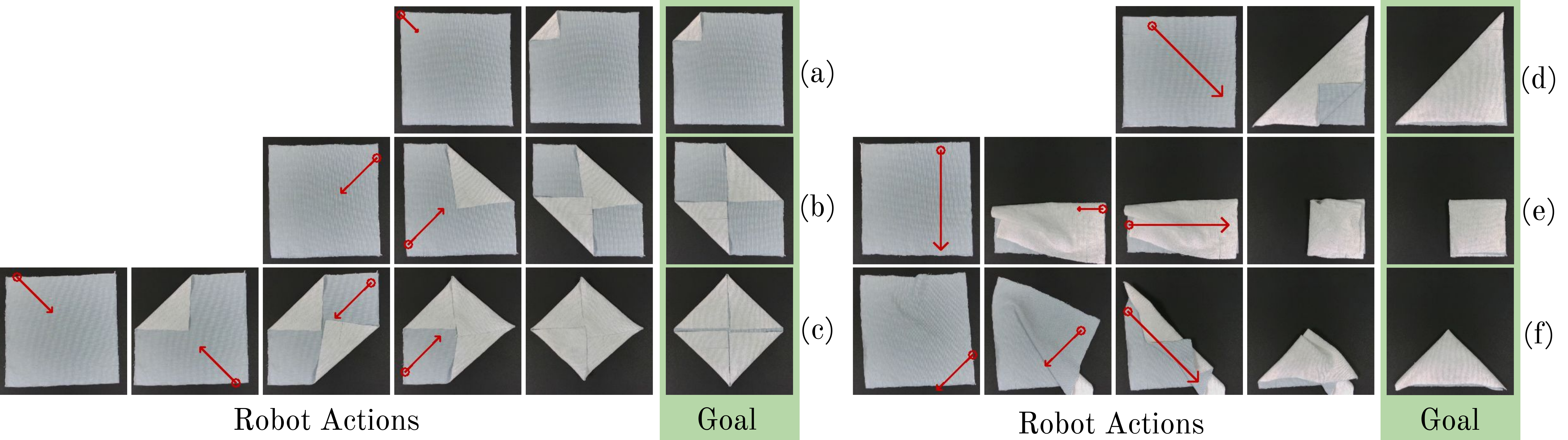}

  \caption{Examples of successful trajectories for each of the six goals. a) Small Inward. b) Double Inward. c) All Corners Inward. d) Triangle. e) Double Straight. f) Double Triangle. Robot actions are visualized as red arrows.
    }
    \label{fig:exampleTrajectories}
\end{figure}

We consider six folding tasks which are unseen during training and executed in the real world. At the start of each trial the fabric is flattened and centered in the workspace. If the goal state is reached, the episode is ended and considered successful. All goals are shown in Figure \ref{fig:exampleTrajectories}. Of the six goals, we consider two challenging compositional folds ((e) and (f)) which involve multiple fabric layers. These tasks are particularly challenging as the fabric is self-occluding, while the goal is provided as a single image.

%===============================================
\subsection{Baselines}
\label{sec:baselines}

We compare our performance to a random action baseline, an ablated version of our method, and prior work by Ganapathi~\etal~\cite{ganapathi2020learning}.
For the random baseline, we employ the same random action policy utilized during data collection. When the fabric is in the unfolded and open, i.e. the initial condition in these experiments, this random behaviour occasionally results in folds from the outer edges of the fabric towards or across the centre.

For the ablation experiment, we evaluate the effect of our scaling technique for achieving discrete distance folds. Instead of performing image scaling to produce the fold distance, for each rotated image, the Q-network instead produces three heatmaps, one for each discrete distance. It is trained using the same procedure, setup and dataset as our approach.

Three of our folding tasks (folds (b), (d) and (f)) are the same as Ganapathi~\etal~\cite{ganapathi2020learning} which enables comparison of the two methods. They evaluate their algorithm on two robots, a Da Vinci Surgical Robot Research System (dVRK) and a YuMi. Our results are more closely comparable to those on the YuMi because the robot is more similar to our Panda system. The dVRK, however, was designed for a very different and more precise application domain.

%===============================================
\subsection{Performance Metrics}
\begin{figure}[t!]
   \centering
   \begin{tabular}{@{}cc@{}}
      \includegraphics[width=0.45\textwidth]{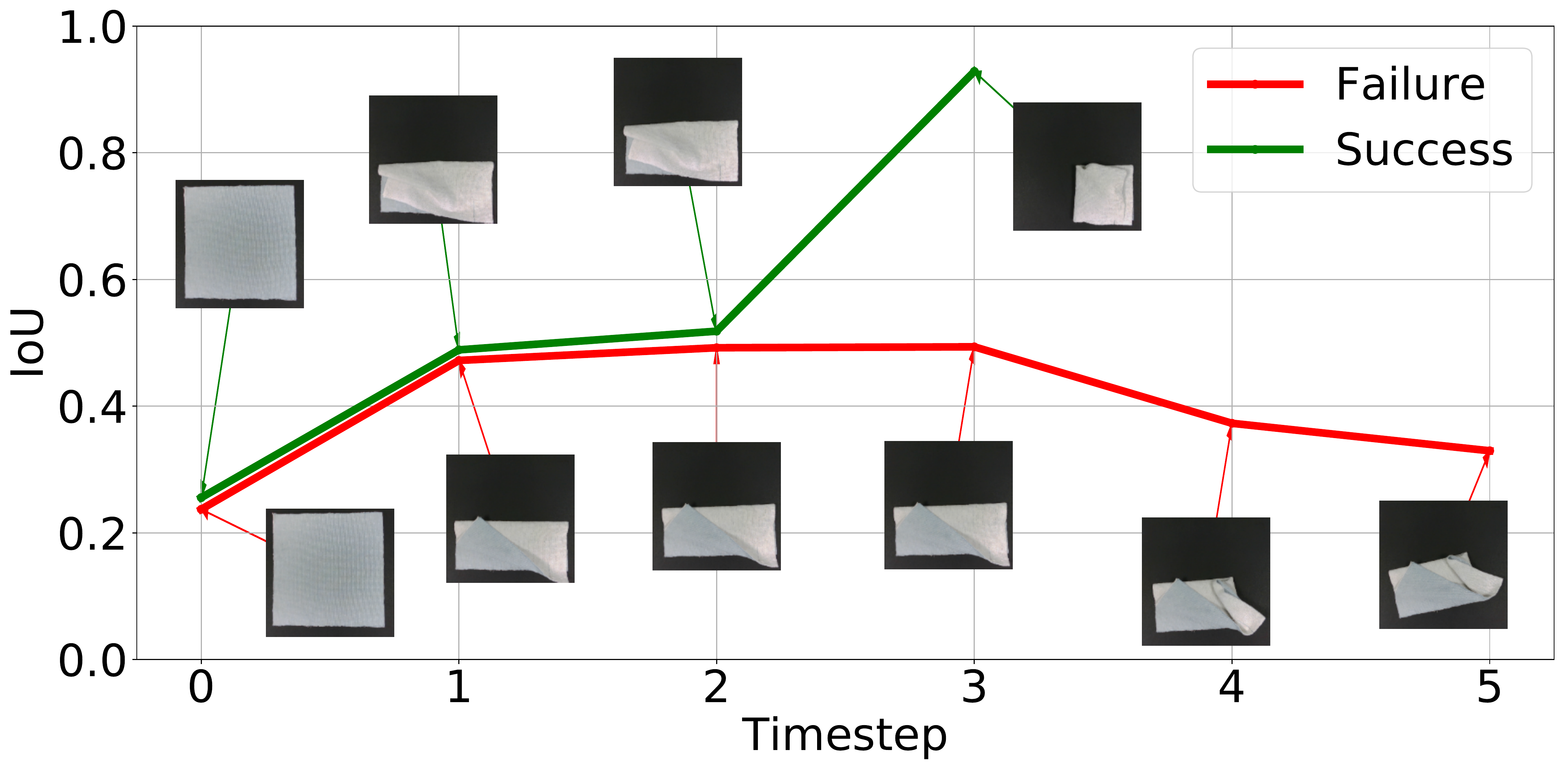} \newline
    & \includegraphics[width=0.45\textwidth]{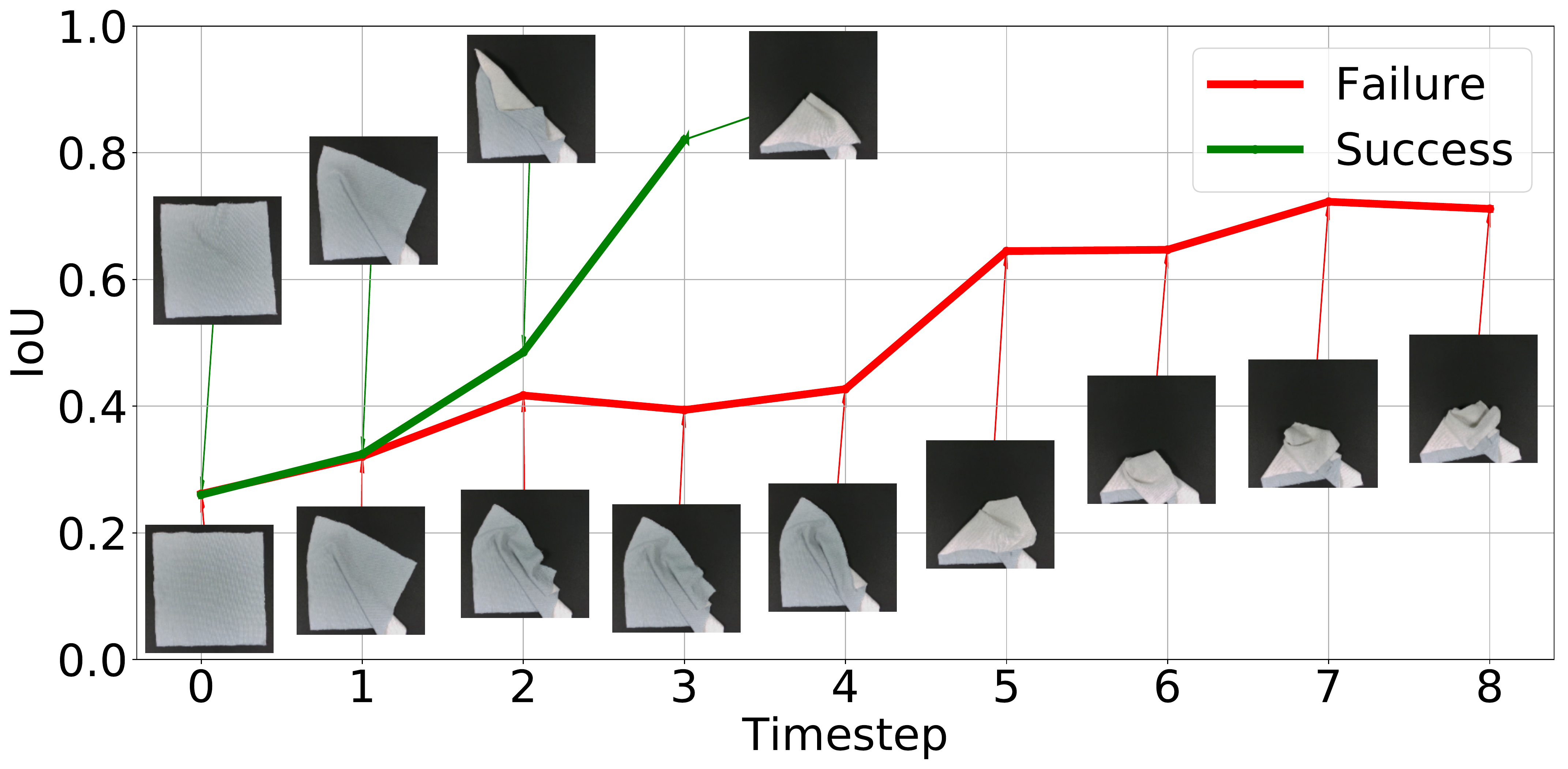} 
   \end{tabular}
   \caption{A comparison of IoU over time between successful and failed for the Double Straight (Left) and Double Triangle (Right) folds.}\label{fig:iouProgression}
\end{figure}

Following Ganapathi~\etal~\cite{ganapathi2020learning} we assess our performance based on fold success rate. The success criteria for a fold is the fabric being ``visually consistent with the target image". It is challenging to apply quantitative metrics such as intersection over union (IoU) or structural similarity between the fabric and goal arrangement because of the deformable, self-occluding nature of the fabric and the limitations observing a 3D phenomenon in 2D images. 
IoU is particularly sensitive to the initial state of the fabric and requires registration of images for accurate comparison, thus we do not use it to determine fold success, however as our initial conditions are similar across trials, it can be used to quantify approximate progress towards goals, as seen in Figure \ref{fig:iouProgression}.

%===============================================
\subsection{Folding Task Results}
Table~\ref{tab:SuccessRate} presents the success rate of our method on the six folding tasks. Our approach demonstrates the learned policy by consistently outperforming the random baseline and succeeding at all folds which do not involve self-occlusion of the fabric in all runs. On the more challenging \textit{double straight} and \textit{double triangle} folds, we achieve 6 and 1 successes out of 10 trials respectively. We visualize successful trajectories performed by our approach, for all goals, in Figure \ref{fig:exampleTrajectories}.

In our approach we account for various fold angles and action distances by discretizing input rotations and scales respectively. In achieving consistent performance across multiple different folding tasks which involve folds of various angles or sizes this approach is validated. For example \textit{small inward} and \textit{single (triangle)} fold require the same action of different distances and the \textit{four corners inward} fold requires multiple folds of different angles, and our method successfully selects the correct angles and distances to achieve the goals.

Without the scaling technique that our method employs, the folding task becomes an inherently harder learning problem, as shown by the ablation experiment in Table \ref{tab:SuccessRate}.

\begin{figure}[tbh]
   \centering
      \includegraphics[width=0.45\textwidth]{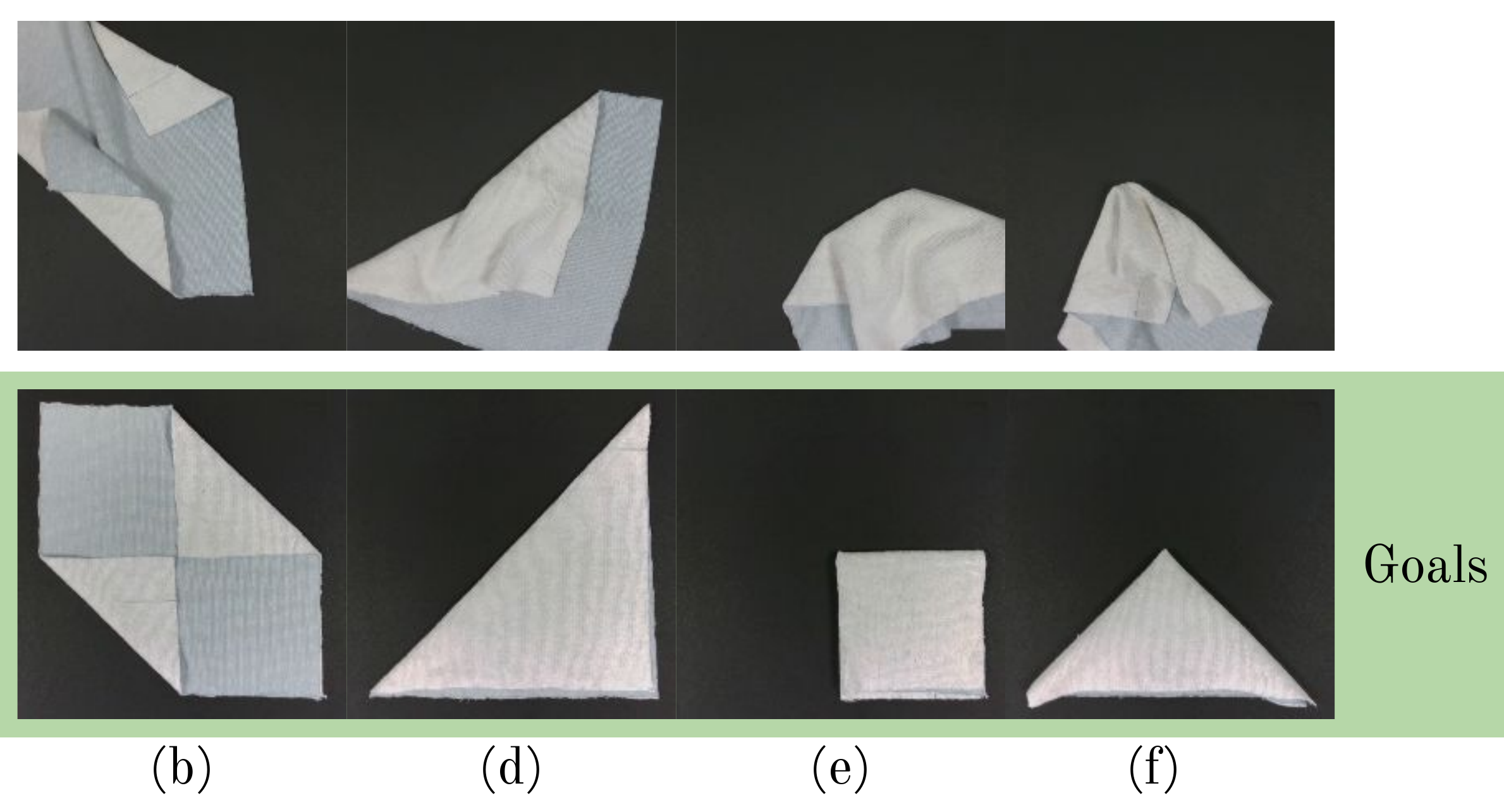}% \newline
   \caption{Examples of failed attempts performed by the ablated baseline approach.
   Despite ``failing" consistently, we note the policy has learned some fabric manipulation ability.}
   \label{fig:ablationfails}
\end{figure}

The ablation baseline approach fails to reach all goals except the \textit{double inward} fold, which it achieves twice. This approach was unable to consistently apply the correct fold distance in the appropriate situations. However, despite failing all other goals consistently, we note that the approach has learned some folding ability, as shown by the failed attempts in Figure \ref{fig:ablationfails}. We note also that this approach does not have the same action-space generalization ability that ours maintains; the model must be trained on a specific discretized action-space, while our approach can generalize to higher resolution action-spaces at test time.

\begin{table}[t!]
\centering
\begin{tabular}{l|ccccc}
\hline

\textbf{Goal}                     & \textbf{\cite{ganapathi2020learning} (YuMi)} & \textbf{\cite{ganapathi2020learning} (dVRK)} & \textbf{Random} & \textbf{Ours (No scale)} &  \textbf{Ours} \\ 
\hline%\hline
\textit{a) Small Inward}        & -                                & -                                & 1            & 0        & 10        \\ 
% \hline
\textit{b) Double Inward}       & 8                             & 9                             & 0            & 2        & 10        \\ 
% \hline
\textit{c) Four Corners Inward} & -                                & -                                & 0            & 0        & 10        \\ 
% \hline
\textit{d) Single (Triangle)}   & 8                             & 9                             & 2            & 0        & 10        \\ 
% \hline
\textit{e) Double Straight}     & -                                & -                                & 0            & 0                  & 6        \\ 
% \hline
\textit{f) Double Triangle}          & 6                       & 8                             & 0            & 0                & 1         \\ 
\hline
\end{tabular}
\vspace{0.75mm}
\caption{The success rate for the six folding tasks of our folding experiments. Each value is the number of successes out of 10 conducted trials. We observe consistently greater performance over the random baseline and comparable performance to previous work on similar folding tasks.}
\label{tab:SuccessRate}
\end{table}

For the more complex goals in our test set (\textit{double straight, double triange}), despite being unable to directly observed the required folds, the system must infer the next best action in the sequence that will achieve the goal image configuration. Occasionally, the policy will make small adjustment actions. In the case of the \textit{double straight} task in Figure~\ref{fig:iouProgression}, the network is sensitive to the overlapping top layer at the bottom right corner of the image. In the successful cases, it is able to reduce the visual appearance of this overlap via small adjustments and proceed to the goal.

However, in the failed case, the policy seems unable to correctly adjust, and therefore does not reach the goal.

In the \textit{double triangle} task shown in Figure~\ref{fig:iouProgression}, the policy chooses to fold the top right corner over in two steps. However, this strategy often fails due to the complexity of folding two layers at once. We also note that, in a real robot setting the grasp quality and reliability must be considered. The challenges of grasping multiple layers of fabric in the \textit{double straight} and \textit{double triangle} folds impacted performance, which is a physical limitation of the gripper.

%===============================================
\subsection{Generalization}
\label{sec:gen}
\begin{figure}[tbh]
  \centering
  \includegraphics[width=0.35\textwidth]{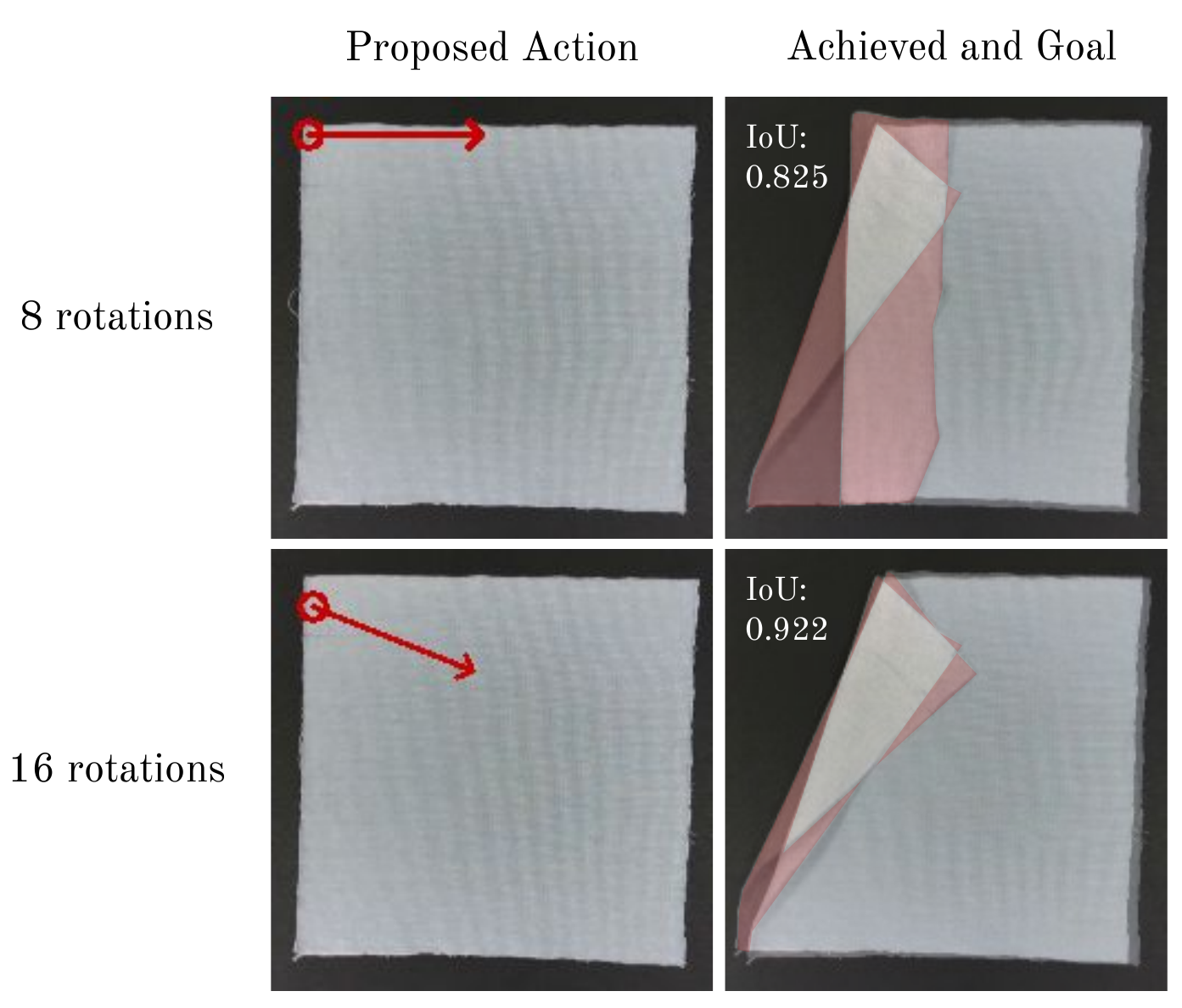}
  \includegraphics[width=0.35\textwidth]{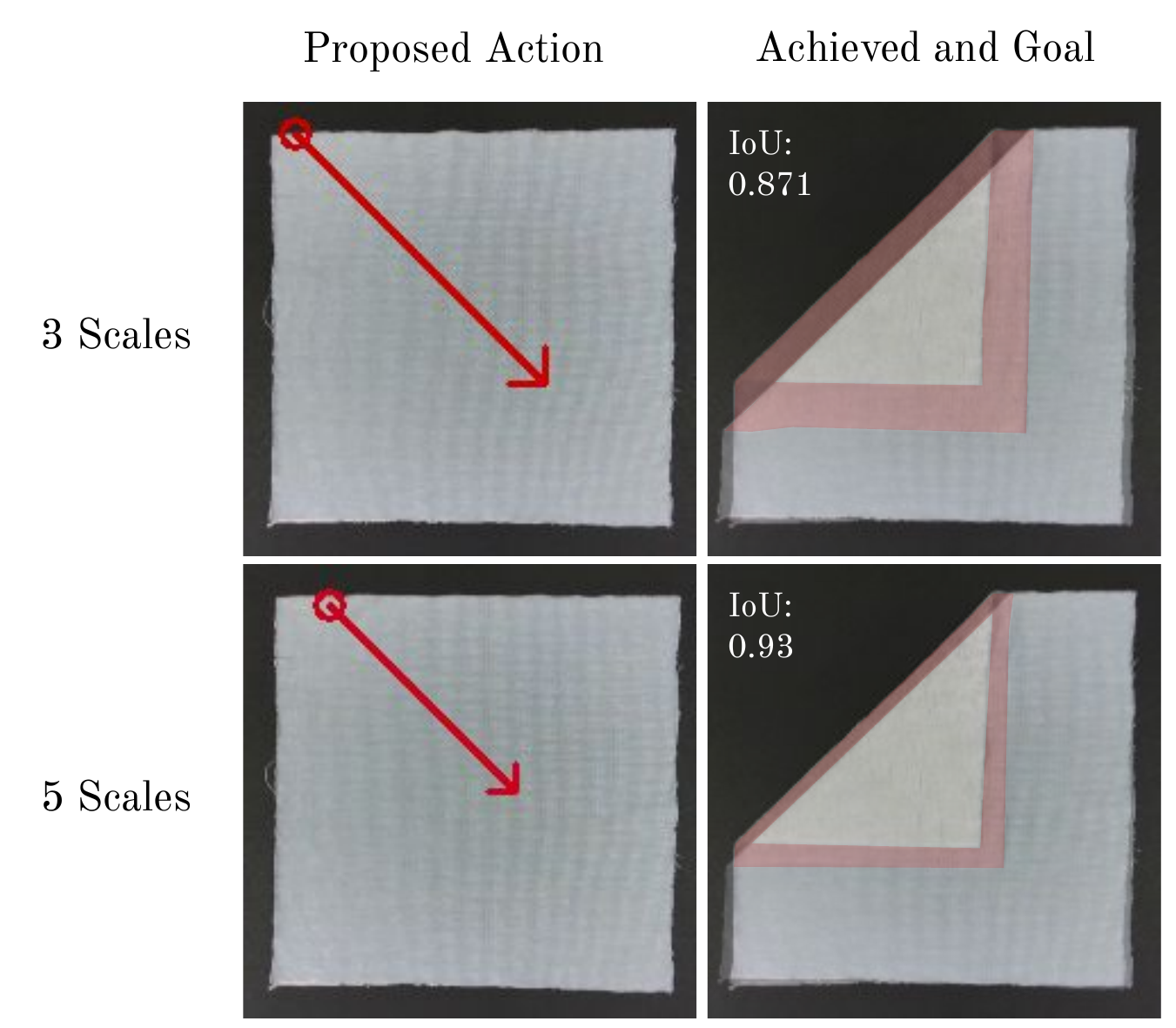}
  \caption{The capability of our model to perform under a higher resolution action-space discretization than experienced during training. Goal is shown overlayed over the state achieved by our approach, with error highlighted in red. Left: we increase the number of angle bins to from 8 to 16.
  Right: We increase number of fold distance bins in the distance discretization from 3 to 5.}
  \label{fig:resolutionDiscretisation}
\end{figure}

The complexity of learning to fold is significantly reduced by discretizing the input into eight possible rotations and 3 possible scales. This set of discretizations enables our method to be sample efficient and is expressive enough to solve a wide range of goals, as shown in Figure \ref{fig:exampleTrajectories}. However, for more complex goals, or when more precision is required, our method can be scaled up to higher resolution discretizations without retraining the model. Our method is able to effectively utilize a larger action space at test time.
Figure~\ref{fig:resolutionDiscretisation} presents the outcomes for two folding tasks which were hand-designed to require an angle and scale not present in the discrete options. The ability to specify an arbitrary discretization resolution at test time without retaining the model demonstrates its generalization ability. In both cases, the model is able to more accurately perform folds it is unable to achieve with the lower resolution action-space used for data collection and training.

\pagebreak
\section{Conclusion}\label{sec:conclusion}
In this work, we present a framework for learning fabric folding directly on a robot in the real world, without requiring human demonstrations or simulated data.

Our offline reinforcement learning approach is able to solve unseen, complex, sequential fabric folding tasks, with only one hour of real data collected in advance.

We achieve this by utilizing a goal-conditioned, fully convolutional network, trained offline with Hindsight Experience Replay. By discretizing rotation and scale, and exploiting the visual proportionality of folding distance, our approach is extremely sample efficient, yet expressive enough to solve complex fabric folding tasks.
Further, we show the accuracy of the behaviors can be increased after training by simply increasing the discretization resolution of the action-space, allowing our approach to solve a wider range of goals.

In future work, we aim to extend our approach and address several limitations. Specifically, we would like to improve in the area of long-horizon sequential goals, and improve the policy's ability to perform toward the partially occluded goals. It would be desirable if the system could detect a successful final state, or whether the configuration is unable to be recovered from.

%===============================================================================

% The maximum paper length is 8 pages excluding references and acknowledgements, and 10 pages including references and acknowledgements

\clearpage
% The acknowledgments are automatically included only in the final version of the paper.
% \acknowledgments{If a paper is accepted, the final camera-ready version will (and probably should) include acknowledgments. All acknowledgments go at the end of the paper, including thanks to reviewers who gave useful comments, to colleagues who contributed to the ideas, and to funding agencies and corporate sponsors that provided financial support.}

%===============================================================================

% no \bibliographystyle is required, since the corl style is automatically used.
\bibliography{references}

\end{document}